\documentclass{article}

\usepackage[final,main]{neurips}

\usepackage[utf8]{inputenc}
\usepackage[T1]{fontenc}
\usepackage{hyperref}
\usepackage{xurl}
\usepackage{amsmath, amssymb, amsthm}
\usepackage{dsfont}
\usepackage{enumitem}
\usepackage{graphicx}
\usepackage{booktabs}
\usepackage{multirow}
\usepackage{tikz}

\DeclareMathOperator*{\argmax}{arg\,max}
\newcommand{\VoI}{\operatorname{VoI}}
\newcommand{\bpost}{\hat{b}}
\newcommand{\E}{\mathbb{E}}
\newcommand{\R}{\mathbb{R}}

\newcommand{\BRop}{\operatorname{BR}}
\newcommand{\BR}{\BRop^{*}}
\newcommand{\BRhat}{\BRop}
\newcommand{\Ind}{\mathds{1}}

\newtheorem{theorem}{Theorem}
\newtheorem{definition}[theorem]{Definition}

\newtheorem{corollary}[theorem]{Corollary}

\title{Residual Expertise Is Not Decision Value}

\author{
    Nidhish Shah \thanks{Equal contribution.} \\
    Independent Researcher \\
    \texttt{shah@nidhish.dev} \\
\And
    Shaurjya Mandal \footnotemark[1] \\
    Harvard University \\
    \texttt{smandal2@mgh.harvard.edu} \\
\And
    Asfandyar Azhar \\
    University of Oxford \\
    \texttt{asfandyar.azhar@st-annes.ox.ac.uk} \\
}

\hypersetup{
    pdftitle={Residual Expertise Is Not Decision Value},
    pdfauthor={Nidhish Shah, Shaurjya Mandal, Asfandyar Azhar}
}

\begin{document}

\maketitle
\setcounter{footnote}{0}

\begin{abstract}
Audits of human--AI complementarity target the wrong estimand.
They ask whether the human knows more than the model; deployment asks whether that knowledge would change the action.
Residual predictive expertise is not residual decision value.
Deployment rewards actions, and a posterior movement that does not cross a reward-induced action boundary changes nothing; a signal can be informative under log-loss while leaving the model-only action optimal.
We formalize the missing quantity as \emph{boundary regret}: the regret of retaining the model-only action after observing the human signal.
In any finite-action Bayesian decision problem, the value of consulting the human equals expected boundary regret, formally verified in Lean 4.
The identity turns when humans help into a geometric question: which decision facets does the human signal move belief across?
Complementarity is therefore reward-relative.
A given human signal can be valuable under one reward matrix and worthless under another, the human's knowledge unchanged; the boundary moved, not the expertise.
The gap is widest precisely where asymmetric costs dominate: high-stakes clinical and operational decisions.
Audits built on residual predictive expertise overstate the value of human review and route scarce review toward cases where the human is informative but the action will not move.
Boundary regret is the quantity to measure.\footnote{Code and Lean~4 proofs: \href{https://github.com/nidhishs/residual-expertise-is-not-decision-value}{\texttt{github.com/nidhishs/residual-expertise-is-not-decision-value}}.}
\end{abstract}

\section{Introduction}
\label{sec:intro}

Human review is often justified by asking whether the human knows something the model does not.
In clinical AI, a radiologist may catch findings that an imaging model misses, and residual-expertise audits can test whether the human signal carries predictive information beyond the model \citep{AlurEtAl2023,AlurRaghavanShah2024}.
But deployment rewards actions, not predictive information.
A radiologist can shift the posterior without changing the admit/discharge decision; only posterior movements that cross the reward-induced action boundary matter in practice.
Residual predictive expertise and residual decision value are not the same estimand.

Complementarity is therefore reward-relative.
The same human signal can matter under one reward matrix and not under another, not because the human's expertise changed, but because the boundary moved.
Human--AI complementarity is not a property of the prediction problem alone.
For deployment, it also depends on the model belief, the human signal, the action set, and the reward matrix.
When audits treat residual predictive expertise as residual decision value, they can spend scarce review budget on cases where the human is informative but operationally irrelevant.

We formalize the missing estimand as \emph{boundary regret}: the regret of retaining the model-only action after observing the human signal.
For model belief $b_x$, model-only action $a_x = \argmax_a r_a \cdot b_x$ (optimal under $b_x$), human-updated belief $b_{x,h} = P(Y \mid x, h)$, and terminal value $V(b) = \max_a r_a \cdot b$,
\begin{equation}
  \BR(x, h) \;=\; V(b_{x,h}) \;-\; r_{a_x} \cdot b_{x,h},
\end{equation}
which is positive exactly when a rival action becomes preferred under $b_{x,h}$.
In any finite-action Bayesian decision problem, the value of observing the human signal equals expected boundary regret (Theorem~\ref{thm:voi-br}), a form of the classical value-of-information identity \citep{RadnerStiglitz1984,DeLaraGossner2020}:
\begin{equation}
  \VoI(H \mid b_x) \;=\; \E_{h \mid x}\bigl[\BR(x, h)\bigr].
\end{equation}
Our contribution is to operationalize this identity as a model-conditioned, reward-local human--AI audit, yielding a per-case, per-reader score for allocating scarce review.
Predictive gain without a reachable boundary crossing has zero decision value for the deployed action set.
We formalize in Lean~4 with Mathlib \citep{DeMouraUllrich2021,MathlibCommunity2020} both the zero-regret boundary condition and the value-of-information identity above.
The paper develops this criterion in four steps:
\begin{itemize}[nosep,leftmargin=1.5em]
  \item We define boundary regret and prove, in Lean~4, that $\VoI(H \mid b_x) = \E_{h \mid x}[\BR(x,h)]$, with a geometric facet-activation criterion that characterizes when boundary regret is strictly positive (Theorems~\ref{thm:voi-br}--\ref{thm:facet-bridge}).
  \item We validate the plug-in estimator $\BRhat$ on synthetic finite-action problems, including a strong-null condition where humans carry substantial predictive signal yet boundary regret is zero everywhere: predictive gain is not sufficient for decision value.
  \item We show on CheXpert that residual predictive expertise and residual decision value separate as a function of reward geometry, exactly as the theory predicts: the same model, the same readers, and the same conditions yield opposite-signed correlations under different rewards.
  \item We show that ex ante $\BRhat$ improves scarce-review allocation over reward-aware margin, reward-agnostic uncertainty, and residual-expertise baselines where actionable mass is dense, and converges to reward-aware margin where it is sparse or absent (Theorem~\ref{thm:facet-bridge}).
\end{itemize}

\section{Boundary Regret}
\label{sec:methodology}

\subsection{Decision Setup}
We work in finite-action Bayesian decision problems \citep{DeGroot1970,Howard1966}.
An instance with features $x \in \mathcal{X}$ has unknown state $y \in \mathcal{Y} = \{1,\ldots,K\}$.
The decision-maker chooses an action $a$ from a finite set $\mathcal{A}$, $|\mathcal{A}| \geq 2$, and receives reward $r_a(y)$ specified by $R \in \R^{|\mathcal{A}| \times K}$.
A \emph{belief} $b \in \Delta^{K-1}$ is a probability vector over states; write $b_x = P(Y \mid x)$ for the model belief at instance $x$.
At any belief, the terminal value is the largest expected reward available from the action set.
The model-only action $a_x$ attains this value under $b_x$.
Each action has a decision region: the beliefs at which that action is optimal.
Formally,
\begin{equation}
    V(b) \;=\; \max_{a \in \mathcal{A}} r_a \cdot b,
    \qquad
    a_x \in \argmax_{a \in \mathcal{A}} r_a \cdot b_x,
    \qquad
    \mathcal{R}_a
    \;=\;
    \{b : r_a \cdot b \geq r_{a'} \cdot b\;\forall a' \in \mathcal{A}\}.
\label{eq:decision-setup}
\end{equation}
Each $\mathcal{R}_a$ is a convex polytope, and the relative boundary $\mathcal{R}_a \cap \mathcal{R}_{a'}$ is a \emph{decision facet}.
The regret of action $a$ at belief $b$ is $\operatorname{regret}(a, b) = V(b) - r_a \cdot b \geq 0$, with equality iff $a$ is optimal.

\subsection{Boundary-Regret Estimand}
After observing the human signal $h$, the model belief $b_x$ updates to $b_{x,h}$.
That update has decision value only if it crosses a decision facet, making some rival action better than $a_x$.

\begin{definition}[Boundary Regret]\label{def:br-instance}
For an instance $x$ with model-only action $a_x$ and human signal $h$, the \emph{instance-level boundary regret} is the regret of retaining $a_x$ after the belief updates from $b_x$ to $b_{x,h}$.
The \emph{ex ante boundary regret} is the expected decision value of the human signal at instance $x$, before $h$ is observed.
Formally,
\begin{equation}
    \BR(x, h) \;=\; V(b_{x,h}) \;-\; r_{a_x} \cdot b_{x,h},
    \qquad
    \BR(x) \;=\; \E_{h \mid x}[\BR(x, h)].
\label{eq:boundary-regret}
\end{equation}
We call $\BR(x,h)$ the \emph{oracle} boundary regret; $\BRhat$ denotes its plug-in estimate (\S\ref{par:estimation}).
\end{definition}

\subsection{Decision-Value Identity}
The value of observing the human signal is exactly expected boundary regret, a form of the classical value-of-information identity \citep{RadnerStiglitz1984,DeLaraGossner2020}.
This rewrites value of information as the loss from keeping the model-only action after the posterior update.

\begin{theorem}[VoI as Expected Boundary Regret]\label{thm:voi-br}
In any finite-action Bayesian decision problem with $a_x$ optimal at $b_x$, where $b_{x,h}$ is the Bayesian posterior update of $b_x$ after observing $h$, the value of observing the human signal equals expected boundary regret:
\begin{equation}
    \VoI(H \mid b_x)
    \;:=\; \E_{h \mid x}\bigl[V(b_{x,h})\bigr] - V(b_x)
    \;=\; \E_{h \mid x}\bigl[\BR(x, h)\bigr]
    \;=\; \BR(x).
\label{eq:voi-br}
\end{equation}
\end{theorem}
\vspace{-1.25em}
\begin{proof}
Use the Bayesian martingale identity: the posterior averages back to the prior.
Therefore the expected reward of any fixed action is unchanged after averaging over the posterior update:
\[
    \E_{h \mid x}[b_{x,h}] = b_x,
    \qquad
    \E_{h \mid x}[r_a \cdot b_{x,h}] = r_a \cdot b_x
    \quad\text{for any fixed action } a.
\]
The left-hand side is the definition of value of information.
Then use the identity above and the optimality of $a_x$ at $b_x$.
Subtract the fixed-action value from the posterior value:
\[
    \E_{h \mid x}[V(b_{x,h})] - V(b_x)
    =
    \E_{h \mid x}[V(b_{x,h})] - \E_{h \mid x}[r_{a_x} \cdot b_{x,h}]
    =
    \E_{h \mid x}[V(b_{x,h}) - r_{a_x} \cdot b_{x,h}].
    \qedhere
\]
\end{proof}

\subsection{Boundary Crossing}
Boundary regret is positive exactly when at least one reachable human-updated posterior leaves the incumbent decision region.
If every reachable posterior stays in $\mathcal{R}_{a_x}$, the human signal may improve prediction, but the model-only action remains optimal.

\begin{theorem}[Facet Activation Bridge]\label{thm:facet-bridge}
For a coherent Bayesian belief/update pair $(b_x,b_{x,h})$, there is positive ex ante boundary regret if and only if some reachable posterior strictly prefers a rival action to the model-only action.
Equivalently, boundary regret is zero if and only if every reachable posterior remains in the model-only action's decision region:
\begin{equation}
    \begin{aligned}
    \BR(x) > 0
    &\quad\Longleftrightarrow\quad
        \exists h,\, a' \neq a_x
        \text{ with } P(h \mid x) > 0
        \text{ and }
        r_{a'} \cdot b_{x,h} > r_{a_x} \cdot b_{x,h},\\
    \BR(x) = 0
    &\quad\Longleftrightarrow\quad
        b_{x,h} \in \mathcal{R}_{a_x}
        \text{ for every } h \text{ with } P(h \mid x) > 0.
    \end{aligned}
\label{eq:facet-activation}
\end{equation}
\end{theorem}
\vspace{-1.25em}
\begin{proof}
By Theorem~\ref{thm:voi-br}, ex ante boundary regret is the expectation of the non-negative instance-level regret.
Thus $\BR(x) > 0$ iff some reachable $h$ has positive instance-level boundary regret.
By definition, this means $V(b_{x,h}) > r_{a_x} \cdot b_{x,h}$, which holds iff some action $a'$ achieves $r_{a'} \cdot b_{x,h} > r_{a_x} \cdot b_{x,h}$.
The converse follows by contraposition: if $a_x$ is optimal at every reachable posterior, then $\BR(x,h) = 0$ for all $h$, hence $\BR(x) = 0$.
\end{proof}

\begin{corollary}[No Crossing, No Decision Value]\label{cor:zero-dv}
If every reachable human-updated posterior $b_{x,h}$ lies in $\mathcal{R}_{a_x}$, then $\VoI(H \mid b_x) = 0$, regardless of how much $H$ improves predictions as measured by log-loss, calibration, or any proper scoring rule.
\end{corollary}

%

\subsection{Deployment and Estimation}
\paragraph{Potential vs.\ Realized Value.}
$\BR(x)$ is an information-theoretic ceiling on the per-instance decision value of review using the specified human signal.
To realize that value, a system still needs correct elicitation, calibrated posterior updating, and operator compliance \citep{Vaccaro2024,BucinaEtAl2021}.

\paragraph{Estimation.}\label{par:estimation}
The identities above are exact for ideal Bayesian beliefs: $b_x$ and $b_{x,h}$ are a coherent prior/posterior update pair.
Empirically, we replace these objects with estimates.
The estimated model belief determines $a_x$; a cross-fitted human-augmented model estimates the updated belief $\bpost_{x,h} \approx P(Y \mid x,h)$.

Thus $\BRhat$ is a plug-in estimate of boundary regret.
For ex ante review allocation, we also estimate $P(h \mid x)$ before human review and compute
\begin{equation}
    \widehat\E_{h \mid x}\!\left[V(\bpost_{x,h}) - r_{a_x} \cdot \bpost_{x,h}\right].
\label{eq:ex-ante-brhat}
\end{equation}
Calibration affects how accurately this plug-in score recovers true boundary regret; Appendix~\ref{app:exp1-miscalibration} reports a targeted binary stress test of systematic belief error.
Section~\ref{sec:experiments} also details how this quantity is estimated for each experiment.

\section{Experiments}
\label{sec:experiments}

We have three sequential experiments.
Section~\ref{sec:exp1} validates the boundary-regret estimator in synthetic settings with closed-form ground truth.
Section~\ref{sec:exp2} tests whether residual predictive expertise and residual decision value diverge on a real human--AI dataset across reward structures.
Section~\ref{sec:exp3} asks whether an ex ante boundary-regret score allocates scarce human review better than standard baselines under asymmetric rewards.

\subsection{Experiment 1: Estimator and Geometry Validation}
\label{sec:exp1}

\paragraph{Question.}
Does the plug-in estimator recover known boundary regret, and remain zero when a safe action dominates the reachable belief region?

\paragraph{Setup.}
We simulate finite-state Bayesian decision problems with known prior, reward matrix, and channels $P(m\mid y), P(h\mid y)$ ($M \perp H \mid Y$).
For each sampled $(y,m,h)$ we compute the exact posteriors $b_m=P(y\mid m)$ and $b_{m,h}=P(y\mid m,h)$, the model-only action $a_m$, and instance-level boundary regret
\[
  \BR(m,h) \;=\; V(b_{m,h}) - r_{a_m}\cdot b_{m,h}.
\]
The estimator does not see the true channels: it estimates $\hat P(m\mid y), \hat P(h\mid y), \hat p(y)$ from finite data via 5-fold cross-fitting with Laplace smoothing, then computes plug-in $\BRhat$ on held-out instances ($N=10{,}000$ per condition).

\paragraph{Conditions.}
The grid varies $G\in\{2,3,5\}$ decision groups and $K\in\{G,2G\}$ latent states, covering the minimal boundary case ($G=2$), the smallest genuinely multi-action case ($G=3$), and a moderate stress test ($G=5$); $K=2G$ checks that estimation does not degrade when multiple latent states share a decision group.
Three roles: C1 is a positive control (actions reward state groups, human signal crosses boundaries), C2 a weak null (prior deep inside one decision region, weak human channel), and C3 a strong null (highly predictive human signal but a safe action dominates every reachable posterior).
For C3 we set $\tau=(1+U)/2$, where $U$ is the maximum reachable posterior mass on any decision group under the exact channels; $\tau>U$ guarantees no reachable posterior can make a non-safe action optimal while the human signal retains its predictive accuracy.

\paragraph{Results.}
Table~\ref{tab:exp1-core} reports metrics averaged across the six configurations; rank correlation is reported only where $\BR$ varies (in both nulls $\BR\equiv 0$, so we report false-positive rate instead).

\begin{table}[htbp]
\centering
\caption{
    Synthetic validation of $\BRhat$ averaged over $G\in\{2,3,5\}$, $K\in\{G,2G\}$.
    C3's $\Delta\mathrm{LL}$ (0.595) exceeds C1's because the human signal is strongly predictive of group membership, yet no posterior crosses the certified threshold.
    $\Delta\mathrm{LL}$: log-loss improvement from $b_m$ to $b_{m,h}$.
}
\label{tab:exp1-core}
\small
\begin{tabular}{@{}llrrrrr@{}}
\toprule
Condition & Role & $\rho(\BRhat,\BR)$ & FPR & $\%\BR{>}0$ & $\%\BRhat{>}0$ & $\Delta$LL \\
\midrule
C1 & Positive control & 0.901        & 0.000 & 37.9 & 37.9 & 0.408 \\
C2 & Weak null        & \text{n/a}   & 0.000 &  0.0 &  0.0 & 0.019 \\
C3 & Strong null      & \text{n/a}   & 0.000 &  0.0 &  0.0 & \textbf{0.595} \\
\bottomrule
\end{tabular}
\end{table}

The estimator separates predictive value from decision value.
In C1, $\BRhat$ ranks true boundary regret well ($\rho=0.901$) across the 37.9\% of instances where crossings occur (per-configuration details in Appendix~\ref{app:exp1-detail}); under the unshifted cross-fitted estimates, FPR is zero in both nulls.
C3 is the critical case: the human signal achieves the highest log-loss improvement of any condition ($\Delta\mathrm{LL}=0.595$), yet $\BRhat$ assigns zero boundary regret to every instance because no posterior crosses the certified threshold.
High predictive accuracy on task-relevant features does not imply decision value when a safe action dominates the reachable belief region.

A paired binary stress test shifts the log-odds of $\hat b_x$ or $\hat b_{x,h}$ by $\pm1$ while holding the simulated cases, fitting protocol, and all other quantities fixed (Appendix~\ref{app:exp1-miscalibration}).
The resulting FPR is component- and geometry-dependent: shifts in $\hat b_x$ yield 49.1--50.9\% FPR in C1, whereas shifts in $\hat b_{x,h}$ yield 23.0--40.4\% FPR in the C3 strong null; the complementary shifts remain zero in these canonical binary cells.


\subsection{Experiment 2: Residual Expertise Does Not Equal Decision Value}
\label{sec:exp2}

\paragraph{Question.}
Does the correlation between residual predictive expertise and boundary regret invert across reward regimes on the same model, readers, and data?

\paragraph{Setup.}
We pair a pretrained DenseNet-121 chest X-ray classifier \citep{Huang2017,Cohen2022} with five board-certified radiologists of the CheXpert reader study \citep{Irvin2019} across five official competition conditions (Atelectasis, Cardiomegaly, Consolidation, Edema, Pleural Effusion), yielding $N=25$ reader--task pairs on 500 frontal radiographs.
For each pair, the ground-truth label $y$ is the majority vote of the \emph{remaining four} readers, which eliminates structural $h$--$y$ contamination.
Model beliefs $b_x$ are temperature-scaled \citep{GuoEtAl2017} per pair on a 30\% calibration split; evaluation uses the remaining 70\% (${\approx}350$ images per pair).
The human-augmented posterior $\bpost_{x,h} \approx P(Y=1 \mid x, h)$ is estimated via 5-fold cross-fitted logistic regression on features $[\operatorname{logit}(b_x),\, h,\, h \cdot \operatorname{logit}(b_x)]$.
Spearman correlations between residual predictive expertise (log-loss gain over $b_x$) and $\BRhat$ are reported across the 25 pairs under each reward regime; label protocol details in Appendix~\ref{app:exp2-protocol}.

\paragraph{Reward Regimes.}
We evaluate three reward matrices $r_{a,y}$, each placing the decision threshold $r_{0,0}/(r_{0,0}+r_{1,1})$~\citep{PaukerKassirer1980} in a distinct region of the model belief distribution.
R1 is the identity ($r_{0,0}=r_{1,1}=1$, errors $0$), threshold $1/2$, in the bulk.
R2 ($r_{0,0}=1$, $r_{1,1}=5$, errors $0$) penalises false negatives five times as heavily as false positives, threshold $1/6$, below the bulk.
R3 mirrors R2 ($r_{0,0}=5$, $r_{1,1}=1$), threshold $5/6$, in the sparse upper tail.
R3 is the critical negative control: the geometry in Theorem~\ref{thm:facet-bridge} motivates the prediction that when the threshold sits in the tail, facet crossings become rare, boundary-regret mass collapses to near zero, and residual predictive expertise decouples from decision value regardless of how much the reader improves the model's log-loss.

\begin{table}[htbp]
\centering
\caption{Boundary-regret diagnostics on $N=25$ reader--task pairs: actionable BR mass collapses from R1 to R3 and BR concentrates near the active facet in all regimes.
  $n_{\mathrm{v}}$: pairs where $\BRhat\not\equiv 0$; $\bar\rho_{\mathrm{facet}}$: mean Spearman between facet distance and $\BRhat$.
}
\label{tab:exp2}
\small
\begin{tabular}{@{}llrrrrr@{}}
\toprule
Reward & Role & $\bar\rho_{\mathrm{LL}}$ & $\rho{>}0/n_{\mathrm{v}}$ & $\bar\rho_{\mathrm{facet}}$ & Top-decile zero-$\BRhat$ & $n_{\mathrm{v}}$ \\
\midrule
R1 (thr.\ $1/2$) & Primary result    & $\mathbf{+0.477}$ & $25/25$ & $-0.599$ & $11.2\%$          & $25$ \\
R2 (thr.\ $1/6$) & Intermediate      & $-0.051$          & $15/25$ & $-0.866$ & $28.3\%$          & $25$ \\
R3 (thr.\ $5/6$) & Negative control  & $-0.112$          & $0/7$   & $-0.221$ & $\mathbf{97.4\%}$ & $7$  \\
\bottomrule
\end{tabular}
\end{table}

\paragraph{Reward-Conditional Dissociation.}
The correlation between residual predictive expertise (log-loss gain over the model alone) and boundary regret inverts with the reward (Table~\ref{tab:exp2}): $\bar\rho_{\mathrm{LL}}=+0.477$ under R1 (25/25 pairs positive), $-0.051$ under R2 (15/25), and $-0.112$ under R3 (0/7; one-sided sign test $p=0.008$, with the remaining 18 of 25 pairs having $\BRhat\equiv 0$).
Same model, same readers, same data; only the reward changes.

The mechanism is geometric.
Under R1, threshold $1/2$ sits in a dense region of model beliefs, so a single reader update routinely crosses a facet and predictive expertise aligns with decision value.
Under R3, the false-positive facet sits in the tail: only 7 of 25 pairs carry any non-zero $\BRhat$, and, on average across pairs, $97.4\%$ of top-disagreement-decile cases (90th percentile of model--reader disagreement $1-b_x[h]$) have $\BRhat=0$.
The R3 correlation $\bar\rho=-0.112$ on these 7 pairs is directionally consistent with the geometric prediction; the collapse of $n_\mathrm{v}$ from 25 to 7 shows that non-zero boundary regret is confined to fewer pairs.
R2 is intermediate ($28.3\%$ top-decile zero-$\BRhat$).
Facet distance correlates negatively with $\BRhat$ under every reward ($-0.599, -0.866, -0.221$ for R1, R2, R3): boundary regret concentrates near the active facet.
R2's stronger concentration follows from the same geometry: with the threshold at $1/6$ nearly all decisions default to treat, so facet crossings tend to come from beliefs already near $1/6$. R1's centrally located threshold produces weaker concentration.
We call this \emph{reward-conditional dissociation}: a consequence of reward geometry, with the fitted beliefs held fixed.

Appendix~\ref{app:exp2-robustness} shows the ordering of mean correlations holds across in- and off-domain pretrained checkpoints (varying the belief geometry) and readers not involved in ground-truth labelling.
As a negative control, Appendix~\ref{app:exp2-cifar10h} evaluates CIFAR-10H: on highly accurate classifiers whose softmax probabilities rarely straddle a reward facet, $\bar\rho$ collapses to near-zero across all rewards ($+0.015$ under R1 vs.\ $+0.477$ on CheXpert R1), consistent with the geometric account in Theorem~\ref{thm:facet-bridge}.

\paragraph{Implication for Predictive Audits.}
A residual-expertise audit in the style of \citet{AlurEtAl2023,AlurRaghavanShah2024} would find substantial predictive value here (mean log-loss gain $0.387$ nats across all 25 pairs).
Under R3, $97.4\%$ of top-disagreement-decile cases have zero estimated boundary regret, on average across pairs.
The \emph{estimand shift} from predictive to decisional complementarity is not a refinement; it changes the criterion for allocating human review.
Section~\ref{sec:exp3} tests this directly.


\subsection{Experiment 3: Review Allocation under Scarce Human Input}
\label{sec:exp3}

\paragraph{Question.}
Given a fixed human-review budget, can an ex ante boundary-regret score identify the cases where reader input will improve realized decision utility, rather than improve prediction or flag model uncertainty?
The prediction is regime-conditional: it should beat reward-aware but human-blind boundary proximity when reader updates can cross reward-relevant decision boundaries, and converge to it when such crossings are sparse or absent.

\paragraph{Setup.}
We use the same 25 reader--task pairs as in Section~\ref{sec:exp2}.
For each case, the model-only belief $b_x$ induces a model-only action $a_x$.
For each review budget $q\in\{5\%,\,10\%,\,20\%,\,50\%\}$, a policy ranks cases using only pre-review information.
Reviewed cases observe the reader response $h$ and take the human-updated action $\argmax_a r_a\cdot\bpost_{x,h}$; unreviewed cases retain $a_x$.
Utility gain is the held-out realized reward improvement over this no-review baseline.
All learned policy scores are 5-fold cross-fitted on the same fold structure used to estimate $\bpost_{x,h}$, so review priority is always computed out of fold.
Appendix~\ref{app:reward-sensitivity} isolates the routing consequences of reward misspecification.

\paragraph{Policies.}
The proposed policy, $\BRhat$, is an ex ante boundary-regret score: it predicts possible reader responses from $x$, predicts the corresponding human-updated posterior, and scores a case by the expected decision value of that update under the current reward.
We compare it against six baselines.
\textbf{Margin} is a reward-aware distance to the active decision facet, our primary comparison baseline: it captures reward geometry without estimating the human signal $P(h\mid x)$.
\textbf{Residual} is a residual-expertise score, the deployable analogue of a predictive audit \citep{AlurEtAl2023}.
\textbf{Entropy} is reward-blind model uncertainty.
These policies span two axes: Margin is reward-aware but human-blind, Residual is human-aware but reward-blind, Entropy is blind to both, and $\BRhat$ is both reward-aware and human-aware.
\textbf{L2D} is the learned full-coverage competitor: like $\BRhat$ it is both reward-aware and human-aware, but reaches that coverage end-to-end through a cross-fitted logistic deferral classifier with reward-sensitive labels $\Ind[R[a_{x,h},y]>R[a_x,y]]$, in the spirit of \citep{MozannarSontag2020,MaoMohri2023}, rather than through the reward's geometry and the reader-channel posterior.
\textbf{Random} is the uninformed baseline.
\textbf{Oracle} is the hindsight upper bound using realized labels.

\paragraph{Regime-Conditional Allocation.}
We evaluate all policies on the same 25 reader--task pairs under each reward matrix.
Table~\ref{tab:exp3} tests whether the dissociation in Section~\ref{sec:exp2} becomes actionable under a scarce review budget.
It does, but only where reward geometry leaves actionable mass reachable: $\BRhat$ beats Margin at every R1 budget, leads or ties across most R2 budgets, and has no systematic advantage in the R3 negative control.
The paired wins, losses, and ties show when $\BRhat$ improves on Margin.

\begin{table}[htbp]
\centering
\caption{Ex ante review allocation under scarce human input.
  Entries report mean held-out utility gain over no review on $N=25$ reader--task pairs.
  $\BRhat$ beats reward-aware Margin where actionable mass is dense (R1), leads or ties where mass is sparse (R2), and has no systematic edge where mass is absent (R3, negative control).
  W/L/T gives paired wins/losses/ties of $\BRhat$ against Margin; bold marks the best deployable policy in each row (Oracle excluded).
}
\label{tab:exp3}
\scriptsize
\setlength{\tabcolsep}{3pt}
\begin{tabular}{@{}ll rrrrr@{\hspace{5pt}}|@{\hspace{5pt}}rr@{\hspace{5pt}}|@{\hspace{5pt}}r@{}}
\toprule
Reward & $q$ & $\BRhat$ & Margin & Entropy & Residual & L2D & Random & Oracle & W/L/T \\
\midrule
R1 & $5\%$  & $\mathbf{+0.0459}$ & $+0.0395$ & $+0.0395$ & $+0.0209$ & $+0.0152$ & $+0.0184$ & $+0.0514$ & $20/2/3$ \\
R1 & $10\%$ & $\mathbf{+0.0883}$ & $+0.0743$ & $+0.0743$ & $+0.0455$ & $+0.0289$ & $+0.0392$ & $+0.1000$ & $24/0/1$ \\
R1 & $20\%$ & $\mathbf{+0.1638}$ & $+0.1438$ & $+0.1438$ & $+0.0915$ & $+0.0697$ & $+0.0761$ & $+0.2000$ & $22/1/2$ \\
R1 & $50\%$ & $\mathbf{+0.3325}$ & $+0.2797$ & $+0.2797$ & $+0.2544$ & $+0.2485$ & $+0.2022$ & $+0.4495$ & $25/0/0$ \\
\midrule
R2 & $5\%$  & $\mathbf{+0.0514}$ & $+0.0490$ & $+0.0429$ & $+0.0330$ & $+0.0477$ & $+0.0313$ & $+0.0514$ & $3/0/22$ \\
R2 & $10\%$ & $\mathbf{+0.0999}$ & $+0.0965$ & $+0.0768$ & $+0.0610$ & $+0.0946$ & $+0.0534$ & $+0.1000$ & $3/0/22$ \\
R2 & $20\%$ & $\mathbf{+0.1914}$ & $+0.1889$ & $+0.1486$ & $+0.1286$ & $+0.1879$ & $+0.1089$ & $+0.2000$ & $5/2/18$ \\
R2 & $50\%$ & $+0.4113$ & $\mathbf{+0.4161}$ & $+0.2955$ & $+0.3069$ & $+0.4134$ & $+0.2664$ & $+0.4865$ & $7/9/9$ \\
\midrule
R3 & $5\%$  & $\mathbf{+0.0029}$ & $+0.0027$ & $+0.0000$ & $+0.0000$ & $-0.0008$ & $+0.0006$ & $+0.0072$ & $1/4/20$ \\
R3 & $10\%$ & $+0.0025$ & $\mathbf{+0.0031}$ & $+0.0000$ & $+0.0000$ & $+0.0008$ & $+0.0007$ & $+0.0072$ & $0/4/21$ \\
R3 & $20\%$ & $+0.0025$ & $\mathbf{+0.0029}$ & $+0.0000$ & $+0.0000$ & $+0.0017$ & $+0.0006$ & $+0.0072$ & $1/4/20$ \\
R3 & $50\%$ & $+0.0027$ & $\mathbf{+0.0029}$ & $+0.0000$ & $+0.0000$ & $+0.0023$ & $+0.0021$ & $+0.0072$ & $1/2/22$ \\
\bottomrule
\end{tabular}
\end{table}

\paragraph{Dense actionable mass (R1).}
$\BRhat$ wins decisively at every budget (sign-test $p<0.001$); decisive-win rates over Margin are $91\%$, $100\%$, $96\%$, $100\%$ (tallies $20/2/3$, $24/0/1$, $22/1/2$, $25/0/0$).
With actionable mass dense in the belief distribution, the reader channel adds signal beyond reward-aware geometric proximity: Margin finds cases near the active facet; $\BRhat$ identifies which of those the reader can move profitably.
$\BRhat$ also achieves higher mean utility than both tested L2D implementations at every budget, with a substantially smaller gap for the shallow-tree classifier (Appendix~\ref{app:l2d-sensitivity}).

\paragraph{Sparse actionable mass (R2).}
The low threshold places most cases on the positive action; sparse boundary-regret mass concentrates near the active facet (exactly where Margin sorts), making Margin a strong baseline. All deployable reward-aware policies obtain much larger absolute gains than under R3.
$\BRhat$ leads or ties Margin at every budget up to $q=20\%$ with at most two paired losses ($3/0/22$, $3/0/22$, $5/2/18$); the $q=50\%$ row is essentially a tie ($7/9/9$, mean difference $0.005$).

\paragraph{Near-zero actionable mass (R3).}
The active facet sits in the sparse upper tail of model beliefs; eighteen of twenty-five pairs have $\BRhat\equiv 0$ across the evaluation set.
Allocation gains are correspondingly tiny: $\BRhat$ and Margin differ by less than $0.0006$ at every budget, with only three to five decisive pairs out of 25.

\paragraph{Human Channel versus Geometric Proximity.}
Reward-aware distance to the active facet is a strong baseline: it already identifies cases near the decision boundary.
The decisive-win rates over Margin under R1 show that $\BRhat$'s estimate of $P(h\mid x)$ and $P(Y\mid x,h)$ add reliable signal on top of that geometric distance where actionable mass is dense.
Where boundary-regret mass is sparse (R2) or near zero (R3), $\BRhat$ and Margin yield similar utility gains.
Residual and Entropy clarify why both axes matter: human information without reward geometry, or model uncertainty without either reward geometry or the reader channel, allocates review less efficiently.
Residual uses the same cross-fitted response and posterior estimates as $\BRhat$ but ranks by predictive information rather than reward-relative regret.
These comparisons are complementary but do not provide an additive decomposition of the utility gain.

\section{Related Work}
\label{sec:related}

Four literatures bear most directly on boundary regret: residual human expertise, deferral and selective prediction, realized human--AI collaboration, and the decision-theoretic value of information.

\paragraph{Residual Human Expertise Audits.}
A line of work asks whether human judgments carry predictive information an algorithm has missed.
\citet{AlurEtAl2023} test for residual expertise by checking whether expert predictions are independent of outcomes conditional on the available features; \citet{AlurRaghavanShah2024} give an algorithmic-indistinguishability account of the same question.
\citet{KleinbergEtAl2018} formalize the selective-labels problem: when outcomes are observed only for cases selected by the human decision-maker, estimating how outcomes depend on both the instance and the human judgment requires accounting for non-random selection.
\citet{GuoEtAl2025} push the question to the feature level via ILIV-SHAP, attributing each input feature's contribution to the agent-complementary VoI of an AI prediction.
What these papers ask is whether humans carry residual predictive information, and which features drive it. We ask something narrower and decision-relevant: whether that information actually changes the reward-optimal action. Residual expertise can sit comfortably alongside zero residual decision value, and a feature-level decomposition does not tell you whether any of its predictive contributions cross a reward-induced decision facet.

\paragraph{Deferral, Selective Prediction, and Reject Options.}
Learning-to-defer methods train systems to choose between acting automatically and handing an instance to a human expert \citep{MadraseEtAl2018,MozannarSontag2020,HemmerEtAl2023,MaoMohri2023}.
Selective-prediction and reject-option methods address the related problem of when a classifier should abstain under a specified loss \citep{RamaswamyEtAl2018,FrancPrusa2023}, and cost-sensitive variants fold asymmetric penalties into the deferral objective \citep{AlvesEtAl2024}.
All of these decide \emph{when} to defer or abstain given a loss, an expert channel, and a collaboration protocol. The question we want to answer comes earlier: whether the human signal can make the model-only action suboptimal under the deployment reward, before any such protocol is on the table.

\paragraph{Realized Human--AI Collaboration.}
A separate body of work studies when teams actually achieve complementarity in deployment.
\citet{Vaccaro2024} synthesize evidence that human--AI combinations often fail to outperform the better standalone agent.
\citet{PengGargKleinberg2025} show that, in binary classification, calibration alone does not prevent a deterministic combination of forecasts from being less accurate than every individual agent.
\citet{SchoefferEtAl2024} show that explanations can change reliance without necessarily helping users identify correct advice, while \citet{GuoEtAl2024reliance} propose a decision-theoretic framework that decomposes observed deployment loss into mis-reliance and signal-perception components.
\citet{McLaughlinSpiess2024} model recommendation design for realized complementarity, and \citet{HemmerEtAl2025} organize the empirical evidence on when complementarity appears.
A related critical strand argues that complementarity also turns on action-set design, evaluation metrics, and interaction design \citep{StraitouriEtAl2025}.
That work is about teams once they are deployed: whether they realize value, what protocols are achievable in principle, how observed loss decomposes after the fact. Our question runs upstream of all of that. Is reward-local decision value in the joint distribution to begin with, before strategy choice, behavioral frictions, interfaces, and incentives have any chance to surface it or hide it?

\paragraph{Value of Information, Decision Geometry, and Decision-Focused Learning.}
Conceptually we are closest to value-of-information theory and decision geometry.
\citet{Blackwell1951} orders information structures by decision value. \citet{RadnerStiglitz1984} and \citet{DeLaraGossner2020} characterize when information value vanishes because posteriors stay inside a single action region, and \citet{KamenicaGentzkow2011} expresses information value through convex value functions and induced decision regions.
Decision-focused learning makes the adjacent algorithmic point that prediction should be trained for downstream decisions rather than generic accuracy \citep{Elmachtoub2022,Wilder2019}, and decision calibration asks when predictions are indistinguishable to downstream best-response rules \citep{Zhao2021}.
In clinical statistics, decision curve analysis \citep{Vickers2006} evaluates predictive models under threshold-encoded reward rather than generic accuracy, plotting net benefit across a range of clinician decision thresholds; that work established the principle that model utility should be assessed under its downstream loss.
Our contribution is to turn this geometry into a human--AI audit.
For each case and reader, we estimate the expected reward gain from using the reader's information rather than retaining the model-only action.
This gives a score for allocating scarce review by decision value rather than predictive expertise.

\section{Conclusion}
\label{sec:conclusion}

The audit question for human--AI complementarity is not whether the human carries information the model lacks but whether that information would change the reward-relevant action.
We named the missing estimand boundary regret and proved that it equals the value of consulting the human in any finite-action Bayesian decision problem; that quantity is strictly positive precisely when some reachable posterior crosses a decision facet.
The plug-in score this identity yields behaves as the bridge predicts: it allocates scarce review where decision-relevant mass is reachable and converges on reward-aware margin where it is not.
Residual predictive expertise alone does not target this reward-relative value of review.

\paragraph{Limitations.}
Boundary regret characterizes the reward-local decision value latent in the joint distribution of $(x, h, y)$.
The results assume a fixed, finite action set and a reward matrix that does not change when the human signal is observed.
Net workflow gains also depend on the elicitation interface, review completion, action adoption, and review cost \citep{Vaccaro2024,BucinaEtAl2021}.
The deployed score is a plug-in estimate of an exact identity; systematic error in either the model belief or human-augmented posterior can change its magnitude and false-positive rate when it moves an estimate across a reward facet.

\paragraph{Broader impact.}
Routing scarce human review to cases that change the reward-relevant action raises operational efficiency in high-stakes settings such as clinical reading and triage.
But the same machinery, fed a misspecified or contested reward, will under-allocate review for populations whose decision facets it does not reach.
Any operational use of $\BRhat$ should pair the score with reward elicitation and a subgroup audit of which populations it removes human review from.

\paragraph{Outlook.}
Boundary regret does not single out the human as the information source.
The same identity prices any candidate channel (a second-opinion reader, an additional diagnostic test, a more expensive feature) by what it would change about the reward-relevant action.
A reward-faithful training signal for learning-to-defer follows: replace accuracy-based deferral labels with the regret of not consulting, and the construction extends to multi-expert and sequential-elicitation policies when the continuation-policy set is finite and fixed, with unchanged rewards and dynamics.
A prospective deployment routing review on $\BRhat$ under operational rewards is the direct test of how much latent decision value a real workflow can extract.
Complementarity is reward-relative. A signal can be informative under any proper scoring rule while leaving the model-only action optimal. The audit that ignores the boundary measures the wrong thing.

\begin{ack}
The authors received no third-party funding for this work and declare no competing interests.
\end{ack}

\bibliographystyle{plainnat}
\bibliography{references}

\clearpage
\appendix

\section{Lean~4 Proof Artifacts}
\label{app:lean}

All formal results used in the paper are verified in Lean~4 using Mathlib.
The development contains five modules: \texttt{Basics.lean}, \texttt{Regret.lean}, \texttt{JensenGap.lean}, \texttt{FacetActivation.lean}, and the umbrella import \texttt{ResidualExpertiseNotDecisionValue.lean}.
Table~\ref{tab:lean-map} maps the paper's claims to Lean declarations.

\begin{table}[ht]
\centering
\caption{Map from paper statements to Lean~4 declarations.}
\label{tab:lean-map}
\small
\begin{tabular}{@{}llll@{}}
\toprule
Paper & Type & Lean declaration & Module \\
\midrule
$V(b) = \max_a r_a \cdot b$ & Def & \texttt{terminalValue} & Regret \\
$\operatorname{regret}(a,b) \geq 0$ & Lem & \texttt{regret\_nonneg} & Regret \\
$\operatorname{regret}=0 \Leftrightarrow a$ optimal & Lem & \texttt{regret\_eq\_zero\_iff} & Regret \\
$\E[b_{\mathrm{post}}]=b$ & Lem & \texttt{prior\_eq\_weighted\_posterior} & Basics \\
Jensen gap $\geq 0$ & Thm & \texttt{jensenGap\_nonneg} & JensenGap \\
Thm~\ref{thm:voi-br} (VoI $=$ $\E[\BR]$) & Thm & \texttt{jensenGap\_eq\_weighted\_regret} & FacetActivation \\
Thm~\ref{thm:facet-bridge} & Thm & \texttt{facet\_activation\_bridge} & FacetActivation \\
\bottomrule
\end{tabular}
\end{table}

\paragraph{Primitives.}
\texttt{Basics.lean} defines \texttt{Belief}, \texttt{ObsKernel}, \texttt{marginalProb}, and \texttt{posterior}, and proves the posterior martingale identity \texttt{prior\_eq\_weighted\_posterior}.
\texttt{Regret.lean} defines \texttt{terminalValue} and \texttt{regret} and proves non-negativity and the zero-regret optimality characterization.
\texttt{JensenGap.lean} defines Jensen gaps over posterior lotteries and proves non-negativity for belief-convex functions.

\paragraph{Paper Theorems.}
\texttt{FacetActivation.lean} contains the two formal claims used in the text.
The weighted-regret identity \texttt{jensenGap\_eq\_weighted\_regret} rewrites the value of information as expected regret of retaining the prior-optimal action after the posterior update.
The bridge theorem \texttt{facet\_activation\_bridge} states that this expected regret is strictly positive exactly when some reachable posterior strictly prefers a rival action.

\section{Synthetic Validation: Per-Configuration Results}
\label{app:exp1-detail}

Table~\ref{tab:exp1-detail} reports per-configuration results across the three estimator-validation conditions of Section~\ref{sec:exp1}.
False-positive rate is zero in every cell: across all 18 configurations the plug-in estimator never assigned positive boundary regret to an instance with $\BR(m,h)=0$.
Under the positive control C1, $\rho(\BRhat,\BR)$ degrades smoothly as the action/state space grows ($G$ from $2$ to $5$, $K$ from $2$ to $10$, giving $0.969 \to 0.745$), reflecting harder channel estimation, not a failure of the identity.
Under the strong null C3, the human signal is by construction strongly predictive ($\Delta\mathrm{LL}$ up to $0.738$), yet the estimator continues to assign zero boundary regret to every instance, the geometric prediction of Theorem~\ref{thm:facet-bridge}.

\begin{table}[htbp]
\centering
\caption{Per-configuration results for synthetic estimator validation. $G$: decision groups; $K$: latent states; $|A|$: actions in the reward matrix. FPR is zero in every cell and is omitted; $\rho$ is omitted when true boundary regret is constant zero (C2, C3).}
\label{tab:exp1-detail}
\small
\begin{tabular}{@{}lrrrrrr@{}}
\toprule
Condition. & $G$ & $K$ & $|A|$ & $\rho(\BRhat,\BR)$ & $\%\BR{>}0$ & $\Delta$LL \\
\midrule
C1 & 2 & 2  & 2 & 0.969      & 35.5 & 0.226 \\
C1 & 2 & 4  & 2 & 0.984      & 36.4 & 0.230 \\
C1 & 3 & 3  & 3 & 0.941      & 37.9 & 0.403 \\
C1 & 3 & 6  & 3 & 0.903      & 38.2 & 0.420 \\
C1 & 5 & 5  & 5 & 0.862      & 39.8 & 0.577 \\
C1 & 5 & 10 & 5 & 0.745      & 39.8 & 0.592 \\
\midrule
C2 & 2 & 2  & 2 & \text{n/a} &  0.0 & 0.006 \\
C2 & 2 & 4  & 2 & \text{n/a} &  0.0 & 0.006 \\
C2 & 3 & 3  & 3 & \text{n/a} &  0.0 & 0.015 \\
C2 & 3 & 6  & 3 & \text{n/a} &  0.0 & 0.019 \\
C2 & 5 & 5  & 5 & \text{n/a} &  0.0 & 0.035 \\
C2 & 5 & 10 & 5 & \text{n/a} &  0.0 & 0.033 \\
\midrule
C3 & 2 & 2  & 3 & \text{n/a} &  0.0 & 0.447 \\
C3 & 2 & 4  & 3 & \text{n/a} &  0.0 & 0.458 \\
C3 & 3 & 3  & 4 & \text{n/a} &  0.0 & 0.592 \\
C3 & 3 & 6  & 4 & \text{n/a} &  0.0 & 0.613 \\
C3 & 5 & 5  & 6 & \text{n/a} &  0.0 & 0.723 \\
C3 & 5 & 10 & 6 & \text{n/a} &  0.0 & 0.738 \\
\bottomrule
\end{tabular}
\end{table}

\subsection{Sensitivity to Systematic Belief Miscalibration}
\label{app:exp1-miscalibration}

The zero FPRs above concern the unshifted cross-fitted estimates.  To isolate the plug-in estimator's sensitivity to systematic belief error, we use the canonical binary cells ($G=K=2$) of C1 and C3.  C1 mixes positive- and zero-regret cases; C3 is the certified strong null with every case in the FPR denominator.  This binary setting also matches the one-vs-rest downstream task without requiring an arbitrary multiclass direction of perturbation.

For a binary fitted belief $b=(b_0,b_1)$, we apply the signed log-odds offset
\[
T_\delta(b)_1 = \frac{e^\delta b_1}{b_0 + e^\delta b_1},
\qquad
T_\delta(b)_0 = \frac{b_0}{b_0 + e^\delta b_1},
\qquad \delta\in\{-1,+1\}.
\]
Thus $\delta=+1$ multiplies the odds of state 1 by $e$, while $\delta=-1$ divides them by $e$.  We shift $\hat b_x$ and $\hat b_{x,h}$ separately, retain the same simulated cases, fitting protocol, oracle boundary regret, and reward tolerance, and recompute only $\BRhat$ and FPR.  The unshifted baseline is $\delta=0$ and is already reported in the main experiment.

\begin{table}[htbp]
\centering
\caption{Paired binary log-odds stress test of plug-in FPR. Each entry is the percentage of oracle-zero cases assigned $\BRhat>0$ after shifting exactly one fitted belief. $n_0$ is the number of oracle-zero cases, using the same tolerance as FPR.}
\label{tab:exp1-miscalibration}
\small
\begin{tabular}{@{}lrrrrrr@{}}
\toprule
Condition & Baseline & $\hat b_x,-1$ & $\hat b_x,+1$ & $\hat b_{x,h},-1$ & $\hat b_{x,h},+1$ & $n_0$ \\
\midrule
C1 & 0.0 & 50.9 & 49.1 & 0.0 & 0.0 & 6,453 \\
C3 & 0.0 & 0.0 & 0.0 & 40.4 & 23.0 & 10,000 \\
\bottomrule
\end{tabular}
\end{table}

The effect is component- and geometry-dependent.  In C1, perturbing the model-only belief changes the incumbent action and creates false positives among its oracle-zero cases, whereas the same posterior-only shifts do not.  In C3, the safe action makes the model-only shifts inert, but perturbing the human-augmented belief can falsely cross the safe-action facet.  These one-at-a-time shifts intentionally need not preserve the Bayesian coherence relation between $b_x$ and $b_{x,h}$: they are an input-level stress test of separately misspecified deployed estimates, not a perturbation of the exact identity.

This is one interpretable intercept-type stress magnitude, not a general robustness guarantee or an estimate of CheXpert calibration drift.  Calibration fitted and evaluated on genuinely independent (or nested) data could reduce correctable systematic bias, but lower calibration error alone does not guarantee lower boundary-regret FPR near a reward facet.  Standard conformal prediction instead targets coverage of prediction sets; it neither calibrates these point posteriors nor directly controls $\Pr(\BRhat>0\mid\BR=0)$.

\section{CheXpert Protocol}
\label{app:exp2-protocol}

\paragraph{Reader Study and Label Construction.}
The five board-certified radiologists (\texttt{bc1}, \texttt{bc2}, \texttt{bc3}, \texttt{bc5}, \texttt{bc7}) provided binary disease-presence ratings under the official CheXpert five-condition protocol (Atelectasis, Cardiomegaly, Consolidation, Edema, Pleural Effusion) on the 500-image frontal-view test set.
For each (condition, ground-truth reader) pair, the held-out label $y$ is computed by \emph{leave-one-out majority vote}: for reader $r_i$, $y = \Ind[\sum_{j \neq i} v_j \geq 3]$, where $v_j \in \{0,1\}$ are the remaining four ground-truth readers' votes.
This construction prevents structural leakage of $h$ into $y$: the test reader is never a voter in their own ground truth.
The three benchmark readers (\texttt{bc4}, \texttt{bc6}, \texttt{bc8}) were not voters in the original study, so their pairs use all five ground-truth votes (equivalent to the official majority).

\paragraph{Model Belief and Calibration.}
Model probabilities $b_x^{\mathrm{raw}}$ are produced by the \texttt{torchxrayvision} DenseNet-121 \citep{Huang2017,Cohen2022} with the standard X-ray preprocessing pipeline (8-bit normalisation, centre crop, $224{\times}224$ resize).
The primary experiment uses the in-domain checkpoint \texttt{densenet121\allowbreak-res224\allowbreak-chex}; the robustness analyses below use all six publicly available checkpoints.
For each (condition, reader) pair we apply per-pair logit-space temperature scaling: a stratified $30\%$ calibration split (seed $42$), with scalar $T$ found by NLL minimisation in $[0.1,10]$, gives the calibrated belief $b_x = \sigma(\operatorname{logit}(b_x^{\mathrm{raw}})/T)$.

\paragraph{Human-Augmented Posterior.}
On the remaining $70\%$ evaluation split, we estimate $\bpost_{x,h} \approx P(Y=1\mid x,h)$ by $5$-fold cross-fitted logistic regression on features $[\operatorname{logit}(b_x),\,h,\,h\cdot\operatorname{logit}(b_x)]$, with folds stratified on $y$ (\texttt{StratifiedKFold}).
The leave-one-out label construction prevents structural leakage of $h$ into $y$; cross-fitting evaluates each case with a posterior model trained on the other folds, so its plug-in $\BRhat$ uses out-of-fold posterior predictions.

\paragraph{Evaluation.}
For each pair, all per-instance quantities ($\BRhat$, log-loss gain, facet distance) are computed on the held-out fold of each instance and aggregated across folds.
Pair-level summaries (Spearman $\rho$ between log-loss gain and $\BRhat$, top-decile zero-mass fraction) are then aggregated across the $25$ pairs.
Pairs with $\BRhat\equiv 0$ across the entire evaluation set are excluded from the $\rho$ computation; we report the count of valid pairs $n_{\mathrm{v}}$ alongside.

\section{Reward-Conditional Dissociation: Robustness}
\label{app:exp2-robustness}

Table~\ref{tab:robustness} extends the dissociation analysis of Section~\ref{sec:exp2} to all six publicly available DenseNet checkpoints and all eight CheXpert radiologists ($6 \times 8 \times 5 = 240$ reader--task pairs).
The ordering R1${>}$R2${\geq}$R3 holds on every checkpoint individually: R1 is positive on every checkpoint ($+0.373$ to $+0.530$), R3 ranges from clearly negative (in-domain, $-0.117$ to $-0.040$) to essentially zero (off-domain NIH, $+0.007$), and R1 strictly dominates R2 throughout, with the gap compressing on off-domain pretraining.
Off-domain pretraining (NIH, PadChest) attenuates the R1 signal and inflates R2, consistent with calibration drift moving belief mass toward the lower R2 facet at $1/6$.
The direction of the dissociation is consistent with the geometric account.

\begin{table}[htbp]
\centering
\caption{Reward-conditional dissociation across $6$ DenseNet checkpoints $\times$ $8$ readers $\times$ $5$ conditions $=$ $240$ reader--task pairs.
  Entries report $\bar\rho_{\mathrm{LL}}$, the mean per-pair Spearman correlation between log-loss gain and $\BRhat$, under each reward (R1: thr.\ $1/2$; R2: thr.\ $1/6$; R3: thr.\ $5/6$).
  Bold marks the aggregate dissociation pattern across all $240$ pairs.
}
\label{tab:robustness}
\small
\begin{tabular}{@{}lrrr@{}}
\toprule
Checkpoint & R1 $\bar\rho_{\mathrm{LL}}$ & R2 $\bar\rho_{\mathrm{LL}}$ & R3 $\bar\rho_{\mathrm{LL}}$ \\
\midrule
\multicolumn{4}{l}{\textit{In-domain pretraining}} \\
\texttt{densenet121-res224-chex}      & $+0.489$ & $-0.008$ & $-0.117$ \\
\texttt{densenet121-res224-mimic\_ch} & $+0.496$ & $+0.197$ & $-0.040$ \\
\texttt{densenet121-res224-mimic\_nb} & $+0.495$ & $+0.210$ & $-0.042$ \\
\texttt{densenet121-res224-all}       & $+0.530$ & $+0.151$ & $-0.082$ \\
\midrule
\multicolumn{4}{l}{\textit{Off-domain pretraining}} \\
\texttt{densenet121-res224-nih}       & $+0.373$ & $+0.313$ & $+0.007$ \\
\texttt{densenet121-res224-pc}        & $+0.435$ & $+0.313$ & $-0.050$ \\
\midrule
All 240 pairs                         & $\mathbf{+0.470}$ & $+0.196$ & $\mathbf{-0.056}$ \\
\bottomrule
\end{tabular}
\end{table}

\section{CIFAR-10H: Negative Control}
\label{app:exp2-cifar10h}

CIFAR-10H \citep{PetersonEtAl2019} evaluates seven pretrained CIFAR-10 \citep{Krizhevsky2009} classifiers \citep{ChenCIFARModels} against 50 human annotators across 10 one-vs-rest binary tasks (3{,}500 reader--task pairs).
Pretrained CIFAR-10 classifiers are highly accurate; their softmax probabilities rarely straddle a decision boundary for a given annotator's image subset.
This motivates CIFAR-10H as a negative control: we expect little actionable boundary-regret mass, consistent with the geometry in Theorem~\ref{thm:facet-bridge}.

\begin{table}[htbp]
\centering
\caption{CIFAR-10H negative control: mean Spearman $\bar\rho$ between log-loss gain and $\BRhat$ across $3{,}500$ reader--task pairs.}
\label{tab:cifar10h}
\small
\begin{tabular}{@{}lrrrr@{}}
\toprule
Reward & $\bar\rho$ (valid) & $\bar\rho$ (all) & $N$ valid & Top-decile $\BRhat=0$ \\
\midrule
R1 (thr.\ $0.500$) & $+0.015$ & $+0.007$ & $1{,}668$ & $96.2\%$ \\
R2 (thr.\ $0.167$) & $+0.044$ & $+0.024$ & $1{,}905$ & $94.5\%$ \\
R3 (thr.\ $0.833$) & $-0.008$ & $-0.004$ & $1{,}632$ & $96.9\%$ \\
\bottomrule
\end{tabular}
\end{table}

The R1 correlation ($\bar\rho=+0.015$) is ${\sim}32{\times}$ smaller than the primary CheXpert R1 correlation ($\bar\rho=+0.477$ across the $25$ in-domain reader--task pairs; Section~\ref{sec:exp2}) and ${\sim}30{\times}$ smaller than the $240$-pair robustness average ($+0.470$; Table~\ref{tab:robustness}).
In this negative control, even highly discordant reader responses usually leave the model-only action optimal under the fitted posteriors: mean zero-$\BRhat$ fractions in the top disagreement decile are $94.5$--$96.9\%$ across rewards.

\section{Review Allocation: Robustness and Baseline Comparisons}

\subsection{Checkpoint and Reader Robustness}
\label{app:exp3-robustness}

Table~\ref{tab:exp3-robustness} extends the allocation experiment of Section~\ref{sec:exp3} to all six DenseNet checkpoints and all eight CheXpert radiologists ($240$ reader--task pairs).
Three patterns reproduce the main result.
Under R1 (dense actionable mass), $\BRhat$ has higher mean utility than Margin at every budget ($p<0.0001$ for $q\geq 10\%$).
Under R2 (sparse but reachable mass), the pooled $240$-pair comparison favors $\BRhat$ over Margin at $q\in\{5\%,10\%,20\%\}$ ($p<0.0001$), compared with small, mostly tied differences on the $25$ primary pairs.
Under R3 (near-zero actionable mass), mean utility differences are small, with no systematic paired advantage in either direction (sign-test $p > 0.07$ at every budget; mean differences $\leq 0.0012$ at $q=5\%$ and $\leq 0.0004$ for $q \geq 10\%$).

\begin{table}[htbp]
\centering
\caption{Allocation robustness across $6$ DenseNet checkpoints $\times$ $8$ readers $\times$ $5$ conditions $=$ $240$ reader--task pairs.
  Entries report mean held-out utility gain over no review.
  W/L/T gives paired wins/losses/ties of $\BRhat$ against Margin; bold marks the best deployable policy in each row (Oracle excluded).
}
\label{tab:exp3-robustness}
\scriptsize
\setlength{\tabcolsep}{3pt}
\begin{tabular}{@{}ll rrrrr@{\hspace{5pt}}|@{\hspace{5pt}}rr@{\hspace{5pt}}|@{\hspace{5pt}}r@{}}
\toprule
Reward & $q$ & $\BRhat$ & Margin & Entropy & Residual & L2D & Random & Oracle & W/L/T \\
\midrule
R1 & $5\%$  & $\mathbf{+0.0407}$ & $+0.0403$ & $+0.0403$ & $+0.0192$ & $+0.0101$ & $+0.0195$ & $+0.0513$ & $106/74/60$ \\
R1 & $10\%$ & $\mathbf{+0.0787}$ & $+0.0763$ & $+0.0763$ & $+0.0413$ & $+0.0247$ & $+0.0393$ & $+0.0996$ & $134/71/35$ \\
R1 & $20\%$ & $\mathbf{+0.1534}$ & $+0.1444$ & $+0.1444$ & $+0.0939$ & $+0.0647$ & $+0.0788$ & $+0.1931$ & $167/51/22$ \\
R1 & $50\%$ & $\mathbf{+0.3160}$ & $+0.2936$ & $+0.2936$ & $+0.2653$ & $+0.2267$ & $+0.1947$ & $+0.4132$ & $198/25/17$ \\
\midrule
R2 & $5\%$  & $\mathbf{+0.0429}$ & $+0.0350$ & $+0.0368$ & $+0.0186$ & $+0.0258$ & $+0.0215$ & $+0.0604$ & $112/25/103$ \\
R2 & $10\%$ & $\mathbf{+0.0826}$ & $+0.0665$ & $+0.0685$ & $+0.0385$ & $+0.0541$ & $+0.0426$ & $+0.1078$ & $121/23/96$ \\
R2 & $20\%$ & $\mathbf{+0.1557}$ & $+0.1252$ & $+0.1276$ & $+0.0870$ & $+0.1123$ & $+0.0849$ & $+0.2013$ & $132/22/86$ \\
R2 & $50\%$ & $\mathbf{+0.3213}$ & $+0.3052$ & $+0.2682$ & $+0.2430$ & $+0.2826$ & $+0.2079$ & $+0.4399$ & $110/98/32$ \\
\midrule
R3 & $5\%$  & $\mathbf{+0.0042}$ & $+0.0030$ & $+0.0001$ & $+0.0008$ & $+0.0025$ & $+0.0002$ & $+0.0115$ & $22/31/187$ \\
R3 & $10\%$ & $\mathbf{+0.0052}$ & $+0.0048$ & $+0.0001$ & $+0.0013$ & $+0.0043$ & $+0.0003$ & $+0.0120$ & $18/30/192$ \\
R3 & $20\%$ & $+0.0055$ & $\mathbf{+0.0056}$ & $+0.0001$ & $+0.0020$ & $+0.0049$ & $+0.0005$ & $+0.0120$ & $13/23/204$ \\
R3 & $50\%$ & $+0.0057$ & $\mathbf{+0.0057}$ & $+0.0003$ & $+0.0025$ & $+0.0050$ & $+0.0017$ & $+0.0120$ & $8/18/214$ \\
\bottomrule
\end{tabular}
\end{table}

\subsection{Learning-to-Defer Classifier Choice}
\label{app:l2d-sensitivity}

To assess whether the L2D comparison depends on classifier choice, we replace the logistic classifier with a shallow decision tree (maximum depth $3$), holding the features, reward-sensitive training target, folds, and allocation protocol fixed.
Both implementations use the same observations as $\BRhat$: each fold trains on $280$ cases and evaluates on $70$ held-out cases.
Table~\ref{tab:l2d-sensitivity} uses the released implementation on the same folds as Table~\ref{tab:exp3}.

\begin{table}[htbp]
\centering
\caption{L2D implementation sensitivity on the $25$ primary reader--task pairs. Entries are mean held-out utility gains over no review.}
\label{tab:l2d-sensitivity}
\small
\begin{tabular}{@{}llrrr@{}}
\toprule
Reward & $q$ & $\BRhat$ & L2D (logistic) & L2D-Tree \\
\midrule
R1 & $5\%$  & $+0.0459$ & $+0.0152$ & $+0.0398$ \\
R1 & $10\%$ & $+0.0883$ & $+0.0289$ & $+0.0800$ \\
R1 & $20\%$ & $+0.1638$ & $+0.0697$ & $+0.1523$ \\
R1 & $50\%$ & $+0.3325$ & $+0.2485$ & $+0.3253$ \\
\midrule
R2 & $5\%$  & $+0.0514$ & $+0.0477$ & $+0.0486$ \\
R2 & $10\%$ & $+0.0999$ & $+0.0946$ & $+0.0917$ \\
R2 & $20\%$ & $+0.1914$ & $+0.1879$ & $+0.1857$ \\
R2 & $50\%$ & $+0.4113$ & $+0.4134$ & $+0.3962$ \\
\midrule
R3 & $5\%$  & $+0.0029$ & $-0.0008$ & $+0.0022$ \\
R3 & $10\%$ & $+0.0025$ & $+0.0008$ & $+0.0019$ \\
R3 & $20\%$ & $+0.0025$ & $+0.0017$ & $+0.0018$ \\
R3 & $50\%$ & $+0.0027$ & $+0.0023$ & $+0.0035$ \\
\bottomrule
\end{tabular}
\end{table}

Under R1, L2D-Tree closes approximately $80$--$91\%$ of the original mean-utility gap, although $\BRhat$ remains numerically higher at every budget.
Under R2, the tree does not uniformly improve on logistic L2D; under R3, differences remain tiny and mixed.
These comparisons concern the tested implementations at this sample size.
A sufficiently expressive, consistent L2D learner targeting the same conditional expected utility can recover the same population ordering.

\section{Reward-Misspecification Sensitivity}
\label{app:reward-sensitivity}

Reward dependence is intrinsic to the decision problem.
To isolate its routing consequences, we rank the same $25$ CheXpert reader--task pairs by $\BRhat$ under R1 or R2, fitting the evidence models once per fold and holding them fixed across the two rankings.
Both allocations use R2 actions, utility, and the same no-review baseline; only the cases selected for review differ.

\begin{table}[htbp]
\centering
\caption{Routing-only reward sensitivity. Both columns report mean total utility under R2, not utility gain. $\Delta$ and W/L/T compare R1-ranked against R2-ranked allocation; overlap is the mean fraction of selected cases shared by the two rankings.}
\label{tab:reward-sensitivity}
\small
\begin{tabular}{@{}lrrrrr@{}}
\toprule
$q$ & R1 ranking & R2 ranking & $\Delta$ & W/L/T & Overlap \\
\midrule
$5\%$  & $0.8073$ & $0.8210$ & $-0.0137$ & $0/17/8$ & $0.000$ \\
$10\%$ & $0.8390$ & $0.8688$ & $-0.0298$ & $1/20/4$ & $0.000$ \\
$20\%$ & $0.8942$ & $0.9608$ & $-0.0666$ & $1/23/1$ & $0.000$ \\
$50\%$ & $0.9837$ & $1.1807$ & $-0.1970$ & $0/25/0$ & $0.345$ \\
\bottomrule
\end{tabular}
\end{table}

R2's fivefold false-negative weighting changes both allocation and realized utility: R1-based routing has lower mean utility at every budget, and the selected sets are disjoint for $q\leq20\%$.
Practitioners can hold the evidence models fixed and report allocation disagreement and utility under each prespecified reward; choosing among contested objectives remains a governance decision.
This retrospective analysis is not subgroup validation or prospective evidence.

\end{document}